\documentclass[11pt]{article}

\usepackage[final]{acl}
\usepackage{booktabs}
\usepackage{times}
\usepackage{latexsym}
\usepackage{amsmath}
\usepackage{amssymb}
\usepackage{multirow} 
\usepackage[T1]{fontenc}
\usepackage[utf8]{inputenc}
\usepackage{microtype}
\usepackage{inconsolata}
\usepackage{graphicx}
\usepackage{listings}
\usepackage{placeins}
\usepackage{pgfplots}
\pgfplotsset{compat=1.18}

\usepackage{pifont}

\usepackage{siunitx}
\title{Estimating Uncertainty from Reasoning: A Large-Scale Study of Multi- and Crosslingual MCQA Performance in LLMs}

\author{
Andrea Alfarano\thanks{Equal contribution; authors listed in alphabetical order.}
\thanks{Work done during an internship at Amazon.}\quad
Andrea Bacciu\footnotemark[1]\textsuperscript{1}\quad
Saab Mansour\textsuperscript{1}\quad
Amin Mantrach\textsuperscript{1}\quad
Marcello Federico\textsuperscript{1} \\
\textsuperscript{1}Amazon \\
\texttt{\{andbac, saabm, mantrach, marcfede\}@amazon.com}
}

\begin{document}
\maketitle
\begin{abstract}
Uncertainty estimation (UE) enables LLM-powered systems to recognize when to abstain, yet existing research has predominantly focused on English.
We present the first large-scale evaluation of UE methods across 22 languages, spanning high-, mid-, and low-resource settings. Using two human-curated Q\&A datasets, we compare open and closed box UE methods (nine in total) across different model sizes and architectures while eliciting long-form reasoning, avoiding LLM-as-a-judge and embedding-based scoring, which can introduce evaluation noise.
We report three main actionable findings. First, we find that prompting models to reason in English while keeping questions in low-resource languages substantially improves UE performance, suggesting that comprehension of low-resource languages is largely intact, and that the reliability bottleneck lies in generation rather than understanding.
Second, prompting models to reason in English closes the UE performance gap between low and high-resource languages, demonstrating that generation language matters more than the question language. Third, the choice of UE method should depend on model scale: at smaller scales, open-box probability-based methods outperform alternatives; at larger scales, closed-box self-verbalized uncertainty becomes superior.
Finally, we provide an analysis of threshold selection for selective prediction, offering guidance on calibrating abstention in multilingual settings.
\end{abstract}

\section{Introduction}
Large Language Models (LLMs) have changed how people access and interact with information, supporting tasks from everyday planning to complex question-answering \citep{bommasani2022opportunitiesrisksfoundationmodels}. Recent research has primarily focused on improving task performance, for example through chain-of-thought prompting \citep{10.5555/3600270.3602070} and instruction tuning \citep{ouyang2022training}, enabling models to solve increasingly complex problems. However, high accuracy alone is  insufficient: downstream systems must recognize when an LLM’s answer is not grounded in its knowledge, enabling them to abstain, defer to humans, or fall back to safer behavior. This has motivated a parallel line of work on uncertainty estimation (UE), which seeks to determine when models know the answer, enabling  LLM-powered systems to recognize and communicate their lack of knowledge \citep{DBLP:conf/iclr/KuhnGF23}.
Existing research on LLM uncertainty has predominantly focused on English, leaving limited evidence on whether UE methods maintain their efficacy in other languages, especially in low-resource settings (\citet{DBLP:conf/iclr/KuhnGF23, kossen2024semantic, Santilli_2025, cecere-etal-2025-monte}, inter alia). The only dedicated multilingual UE study we are aware of \citep{xue-etal-2025-mlingconf} evaluates just three methods on five languages, relies on machine-translated data without human post-editing, and uses a multiple-choice, short-answer format rather than open-ended generation. Their dataset yields a median answer length of just one word, meaning this setup primarily evaluates how language affects question comprehension while offering limited evidence about uncertainty during longer, more linguistically rich generation.
Establishing multilingual trustworthiness through UE is also methodologically challenging because standard metrics such as AUROC rely on ground truth labels. When ground truth is approximated via LLM-as-judge, BERTScore, or n-gram overlap, these proxies introduce noise that can misrank uncertainty methods \citep{Santilli_2025}. In multilingual settings, this problem is compounded: neural judges may behave inconsistently across languages, introducing language-dependent bias into UE comparisons. As a result, existing English-centric evaluations do not yet provide decisive evidence that UE methods transfer reliably to low-resource languages and long-form generation.
To address these challenges, we evaluate nine UE methods on two human-curated QA datasets across 22 languages, covering high-, mid-, and low-resource settings and spanning diverse language families and resource levels. We elicit long-form, language-rich reasoning (on average 150 words) while preserving objective correctness from QA labels, thereby avoiding LLM-as-judge and embedding-based proxies.

Our results reveal three key patterns: (1) UE method performance varies largely across language families, consistent with effects of resource availability and linguistic distance; (2) generation language has a larger impact on UE than question language, a factor largely overlooked by prior work; and (3) while the closed-box Self Verbalized method achieves the best overall performance, sampling-based closed-box methods degrade substantially on low-resource languages, whereas open-box probability-based methods maintain more consistent performance across all language resource levels.

We make the following contributions: (i) we propose a label-grounded framework for evaluating UE on open-ended multilingual generation without model-based correctness metrics; (ii) a large-scale study of UE across 9 models and 22 languages; and (iii) practical guidance for deploying uncertainty-aware systems in multilingual settings. Specifically, we address: (RQ1) which UE methods are robust across models and languages, (RQ2) how model scale affects UE efficacy, (RQ3) how the language of reasoning shapes UE quality, (RQ4) UE robustness under cross-lingual answer settings, and (RQ5) threshold selection strategies for selective prediction.

\section{Related Work and Background}
\label{sec:related-work}

Uncertainty estimation (UE) is critical for deploying large language models (LLMs) in high-stakes environments, enabling systems to flag unreliable predictions. Existing literature is commonly categorized by the degree of model access available: \emph{open-box methods}, which utilize internal model representations, and \emph{closed-box methods}, which rely solely on textual outputs.

\subsection{Open-Box Methods}
Open-box approaches derive uncertainty directly from model internals, such as hidden states or token probabilities. 

Inference method approaches require no additional training and aggregate token-level logits into sequence-level scores. Standard metrics include length-normalized log-probabilities, the geometric mean of token probabilities, and minimum token probability \citep{fomicheva-etal-2020-unsupervised, vazhentsev2023efficient}. Inference-only methods remain the standard for open-weights models due to their computational efficiency and lack of supervision requirements.

\subsection{Closed-Box Methods}
Closed-box techniques are essential when access to logits or gradients is restricted (e.g., commercial APIs). These methods rely exclusively on the generated text.

\paragraph{Self verbalized uncertainty}
This approach prompts the LLM to explicitly state its confidence as a score or linguistic expression \citep{kadavath2022language, tian-etal-2023-just}. While accessible, self-verbalization is frequently miscalibrated and often struggles to discriminate correctness effectively compared to logit-based baselines \citep{xiongcan, groot-valdenegro-toro-2024-overconfidence}.

\paragraph{Consistency and Sampling Strategies}
Consistency-based methods rely on the intuition that correct reasoning is stable across stochastic samples. \emph{Semantic Entropy} and its variants \citep{DBLP:conf/iclr/KuhnGF23, farquhar2024detecting, cecere-etal-2025-monte} cluster multiple generations by meaning and compute entropy over these clusters. Other works extend this method using graph theory, modeling generations as nodes in a similarity graph to derive uncertainty from structural properties \citep{lin2024generating, vashurin-etal-2025-benchmarking}.

\subsection{Evaluation and Multilingual Challenges}
UE performance is typically evaluated via \emph{discrimination} AUROC but reliable evaluation remains a challenge.

\paragraph{Ground Truth Matching}
\citet{Santilli_2025} demonstrate that automated ground truth
matching like ROUGE or BERTScore correlate strongly with superficial
features (e.g., length) rather than factual accuracy, inflating
reported UE performance. To address this, we avoid automated ground
truth matching and rely on exact matching of multiple-choice answers
against human-curated labels. Prior work has also used MCQA exact
matching to evaluate model reliability.
For example,
\citet{madhusudhan-etal-2025-llms} study whether models select an
explicit ``I Don't Know'' option as a binary abstention behavior. Our framework differs in three ways: we evaluate continuous-valued
uncertainty signals (rather than binary abstention), extract them
from long-form reasoning traces (rather than the answer choice
itself), and operate across 22 languages (rather than English only).
We estimate uncertainty from reasoning generated \emph{before} the answer, following the inference paradigm of modern reasoning LLMs where the reasoning trace is part of the computation that produces the prediction. We do not consider explanations elicited after the answer, as such post-hoc rationalizations are known to be unfaithful to the underlying computation \citep{atanasova-etal-2023-faithfulness}.

\paragraph{Multilingual Uncertainty Estimation}
While UE is well-studied in English, multilingual evaluation remains underexplored. To the best of our knowledge, \citet{xue-etal-2025-mlingconf} present the only dedicated multilingual UE study, introducing \emph{MlingConf} to benchmark three uncertainty methods across five languages. However, their evaluation relies on machine-translated data without human post-editing and employs a short-answer format with a median response length of just one word. This setup primarily assesses how language affects question comprehension, offering limited insight into uncertainty during extended generation. Our benchmark addresses this gap by eliciting long-form reasoning in the target language while preserving objective correctness labels.

\begin{figure*}[ht!]
 \centering
 \includegraphics[width=0.93\linewidth]{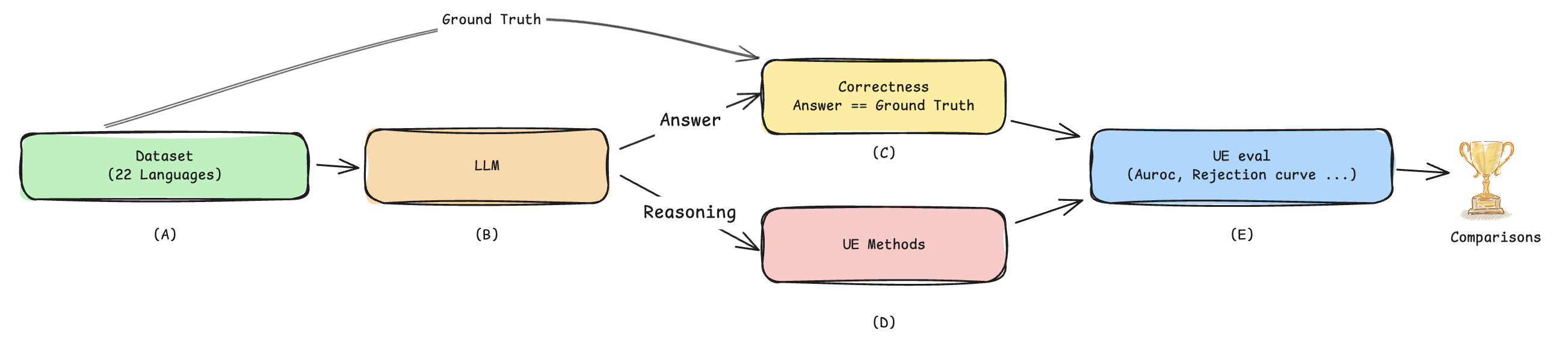}
 \caption{Evaluation pipeline. Unlike prior work that estimates uncertainty from predicted labels in English-only settings, our framework (1) extracts uncertainty from the model's \emph{reasoning trace} and (2) evaluates across 22 languages. The pipeline proceeds as: multilingual prompts (A) → LLM generation (B) → correctness evaluation against ground truth (C) → uncertainty estimation on reasoning (D) → method comparison via AUROC (E).}
 \label{fig:methodology-setup}
\end{figure*}

\section{Methodology}
\label{sec:methodology}
Our goal is to evaluate LLM UE methods across multiple languages, including low-resource ones, while supporting long text generation and using model-free metrics to maintain an unbiased setup.

To achieve this, we leverage human-curated Multiple-Choice Question Answer (MCQA) datasets where the set of possible labels is a fixed set of options (such as A, B, C, and D).
MCQA labels can be interpreted without relying on string-matching, LLM-as-a-judge, or embedding similarity techniques, thus providing ground truth without any approximation or language bias.
MCQA labels are too short to derive uncertainty estimates from the model; for this reason, we apply UE methods to the LLM's reasoning text, similarly to the approach of \citet{podolak2025read}.

Specifically, we prompt the model to generate a reasoning explanation before producing its answer (e.g., Chain of Thoughts \citep{10.5555/3600270.3602070}).
We then apply uncertainty estimation methods to this reasoning text (on average 150 words), while using exact matching against MCQA labels to determine answer correctness.
This allows us to retain the reliability of MCQA labels as ground truth while extending UE evaluation to long text generation.
We deliberately choose MCQA over open-ended QA datasets for two reasons: (i) open-ended answers are extremely short factoids (median 1--2 words), so they do not provide longer text for UE than MCQA labels do, and (ii) their evaluation does not ensure a language- and correctness-unbiased setup. A detailed justification is provided in Appendix~\ref{app:why-not-open-ended}.
Finally, we evaluate the performance of each UE method using the Area Under the Receiver Operating Characteristic (AUROC) curve, which is the standard metric in the UE literature for measuring how well uncertainty scores correlate with answer correctness.
Statistical significance is assessed with DeLong tests on the ROC curves (paired when methods are compared on identical samples, unpaired across model scales), applying Benjamini--Hochberg FDR correction across languages; per-language results are reported in Appendix~\ref{app:significance}.
The complete process is illustrated in Figure \ref{fig:methodology-setup}.

\textbf{Datasets:} We use two human-curated MCQA datasets, namely Global-MMLU \citep{singh-etal-2025-global} (with 4 possible choices) and MMLU-ProX \cite{xuan-etal-2025-mmlu} (with 10 possible choices). We focus on the 22 languages that are common to both datasets (the complete list of languages is available in Appendix \ref{appendix-languages}). These two datasets are parallel, meaning that every question-answer pair is available in every language, making our multilingual evaluation comparable across languages. These two datasets cover a broad set of categories: Global-MMLU contains 6 categories, while MMLU-ProX contains 14 categories (see Appendix \ref{app:question-answering-categories}).
To ensure balanced representation, we sample 100 questions per category from each dataset, yielding 600 questions from Global-MMLU and 1,400 from MMLU-ProX, for a total of 2,000 unique questions. Since both datasets are available in all 22 languages, this results in 44,000 language-specific instances.
We further verify that the MCQA labels are uniformly distributed in both datasets, ruling out positional bias (see Appendix~\ref{app:label-distribution}).

\begin{figure}
    \centering
    \includegraphics[width=\linewidth]{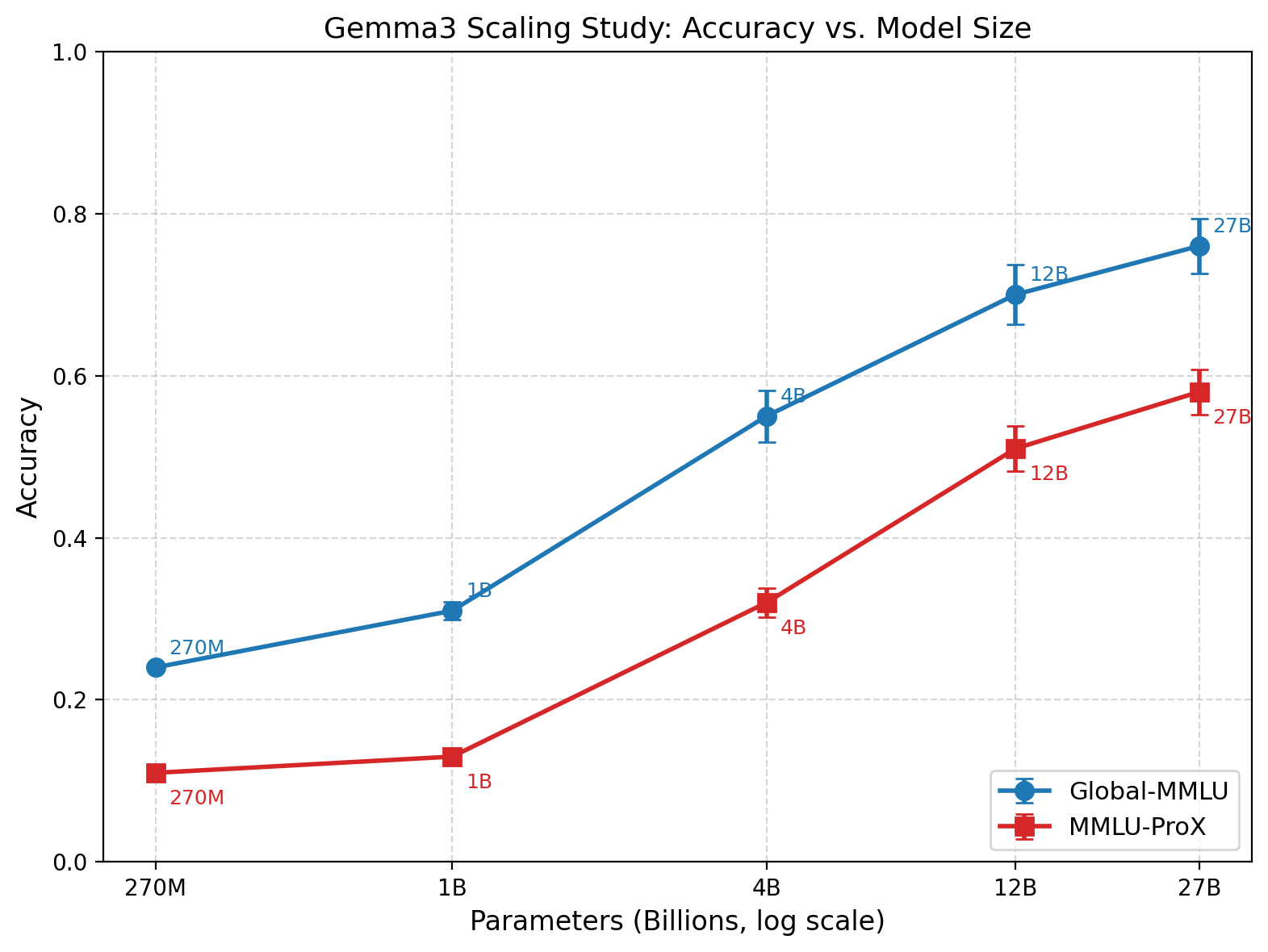}
    \caption{Parameter scaling study on MCQA accuracy on both datasets (mean $\pm$ 95\% CI) over 22 languages using Gemma 3. Similar trends are observed in Qwen models.}
    \label{fig:mcqa-accuracy-scaling}
\end{figure}

\textbf{Models:} Our evaluation spans 9 models from both closed-source and open-source families. For closed-source models, we employ Claude Sonnet 4.5 \citep{anthropic2025claudesonnet45} - the state-of-the-art model at the time of writing. For open-source models, we consider instruction-tuned models including five Gemma3 models (0.27B, 1B, 4B, 12B, 27B) \citep{team2025gemma} and three Qwen3 models (4B, 30B-A3B, 235B-A22B) \citep{yang2025qwen3technicalreport}, encompassing both dense and Mixture of Experts (MoE) architectures. We report a scaling study on the Q\&A accuracy in Figure \ref{fig:mcqa-accuracy-scaling}.
In our experimentation, we applied the quantization only to Qwen3 235B-A22B to 4-bit due to hardware constraints (see our Hardware Infrastructure in Appendix \ref{app:hardware-infrastructure}).
In Appendix \ref{app:quantization}, we investigate whether quantization impacts multilingual UE performance.
We selected Qwen3-30B for this analysis as it is the largest model that fits in memory at full precision. Results show no statistically significant effect when quantizing to 8-bit and 4-bit.

\textbf{Uncertainty Estimation Methods:}
Following the taxonomy in Section~\ref{sec:related-work}, we evaluate nine methods, including both open-box methods (with access to model internal states) and closed-box methods.
Open-box probability baselines include Mean Token Entropy (Token Entropy) \citep{fomicheva-etal-2020-unsupervised}, Maximum Sequence Probability (Max Prob) \citep{vashurin-etal-2025-benchmarking}, and Self Certainty \citep{kang2025scalable}.
Closed-box methods include Self-verbalized uncertainty \citep{tian-etal-2023-just}, multi-sample disagreement based on lexical
overlap as Lexical Similarity (ROUGE-L) \citep{fomicheva-etal-2020-unsupervised,DBLP:conf/iclr/KuhnGF23}, Semantic Entropy with Jaccard-based clustering \citep{DBLP:conf/iclr/KuhnGF23}, and graph-based dispersion (EigValLaplacian\_Jaccard, DegMat\_Jaccard, Eccentricity\_NLI\_Jaccard) \citep{lin2024generating}.
More detailed method definitions are available in Appendix~\ref{sec:ue-methods}.
All prompts used are listed in Appendix~\ref{app:prompts}. All UE methods are implemented through LM-Polygraph \citep{fadeeva-etal-2023-lm,vashurin-etal-2025-benchmarking}.

\section{Results And Discussion} \label{sec:results}
\paragraph{RQ1: Which UE methods are robust across models and languages?} \label{sec:rq3-methods}
\begin{figure*}
 \centering
 \includegraphics[width=\linewidth]{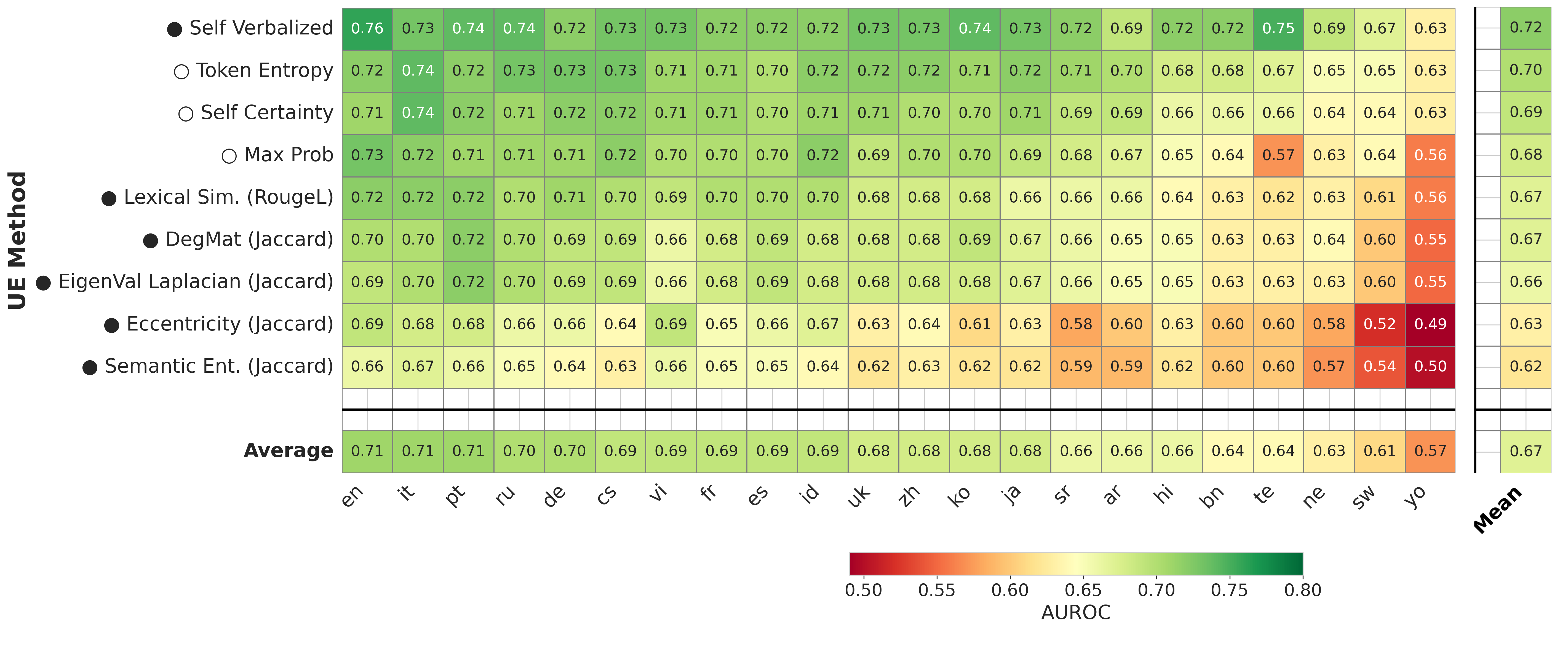}
\caption{Per-language AUROC for nine UE methods across 22 languages on the dense Gemma3-27B and MoE Qwen3-235B-A22B models. Languages are sorted by mean AUROC (median sorting produces identical rankings). Closed-box UE methods are marked with a black circle while open-box are marked with a white circle. In the left block we average across models, in the middle block across languages, while in the right block we average across languages and models. Remarkably, the closed-box method \textit{Self Verbalized} achieves top performance, rivaling open-box methods such as \textit{Token Entropy} and \textit{Self Certainty}—suggesting that effective UE is achievable without access to token-level probabilities.}
\label{fig:method_language_heatmap}
\end{figure*}

To address this question, Figure~\ref{fig:method_language_heatmap} compares nine UE methods across 22 languages and two open-source models (Gemma3-27B and Qwen3-235B). We excluded closed-source models to ensure a fair comparison between open-box and closed-box approaches.

The best-performing method is ``Self Verbalized'', which achieves an average AUROC of 0.72 and demonstrates consistent performance across all languages and models. The second-ranked method is the open-box ``Token Entropy'' with an AUROC of 0.71, followed by ``Self Certainty'' at 0.70. The open-box ``Max Prob'' performs reasonably well on average (0.68) but degrades substantially on low-resource languages, dropping to 0.56 on Yoruba and 0.57 on Telugu.

Sampling-based methods maintain reasonable performance on high-resource languages but fail on low-resource ones, with Eccentricity and Semantic Entropy dropping to near-random performance on Yoruba (0.49 and 0.50 AUROC). These methods generate multiple reasoning traces and use their diversity as an uncertainty signal—the intuition being that confident models produce consistent outputs while uncertain models vary.
To understand why this fails, we analyze the diversity of reasoning traces across all 22 languages using pairwise Jaccard distance. For high-resource languages, the expected pattern holds: incorrect predictions produce more diverse reasoning than correct ones (English: 0.72 vs.\ 0.63, $p < 10^{-6}$; German: 0.73 vs.\ 0.65; Spanish: 0.74 vs.\ 0.63; French: 0.69 vs.\ 0.64), with gaps of 0.08--0.11. For low-resource languages, this gap collapses to $\leq$0.03: Yoruba (0.93 vs.\ 0.93), Swahili (0.93 vs.\ 0.92), Bengali (0.83 vs.\ 0.80).
LLMs generate highly variable reasoning regardless of correctness, leaving no diversity signal to exploit.
Self-verbalized approaches, by contrast, leverage meta-cognitive capabilities \citep{steyvers2026metacognition, xu2025confronting} that transfer more robustly across languages.
The Self Verbalized signal is also robust to the elicitation wording: across three prompt phrasings (Appendix~\ref{app:prompt-ablation}), error-detection AUROC remains stable (0.72--0.76) with no statistically significant difference, confirming the ranking is not an artifact of a single prompt formulation.

\paragraph{RQ2: How does model scale affect multilingual UE method efficacy?}
We investigate the effect of model parameters scale on UE performance, we evaluate the top-2 open-box (Token Entropy, Self Certainty) and top-2 closed-box (Self Verbalized, Lexical Sim.) methods (selected for figure clarity) that scale best across all languages (see RQ1); for all other methods, see Appendix \ref{app:model_scale_all_ue}. In Figure~\ref{fig:ue-method-scaling}, we computed the average across all the languages and averaging across datasets. We select Qwen3 as it offers the broadest parameter range among the open-source models in our study from 4B to 235B. We observe a clear divergence in scaling behavior.
Token Entropy changes only modestly across the evaluated scales, with overlapping confidence intervals. Self Verbalized, by contrast, improves substantially with scale (from 0.65$\rightarrow$0.66$\rightarrow$0.77 AUROC): the difference between 4B and 30B is not significant, while the gain at 235B is significant in 22 of 22 languages on Global-MMLU (unpaired DeLong tests, BH-FDR corrected, mean gap $+0.153$ AUROC; Appendix~\ref{app:significance}). At 4B and 30B scale, Token Entropy significantly outperforms Self Verbalized (19 of 22 languages at 30B, paired DeLong tests; Appendix~\ref{app:significance}), but this relationship reverses at 235B where Self Verbalized achieves the highest AUROC among all methods.
These findings have practical implications: at smaller scales, open-box methods such as Token Entropy provide the most reliable uncertainty signal; at larger scales, Self Verbalized surpasses all alternatives by a significant margin, suggesting that the quality of explicit self-assessment improves substantially at larger model scales, consistent with prior findings on self-knowledge in LLMs \citep{kadavath2022language, steyvers2026metacognition}.
However, training pipelines are not uniform across model sizes within the same family: the Qwen3 Technical Report \citep{yang2025qwen3technicalreport} discloses that smaller Qwen3 variants are distilled from larger teacher models (Qwen3-235B-A22B), so part of the observed gap may also reflect that self-verbalization capabilities are sensitive to training procedures.

\begin{figure}
    \centering
    \includegraphics[width=0.9\linewidth]{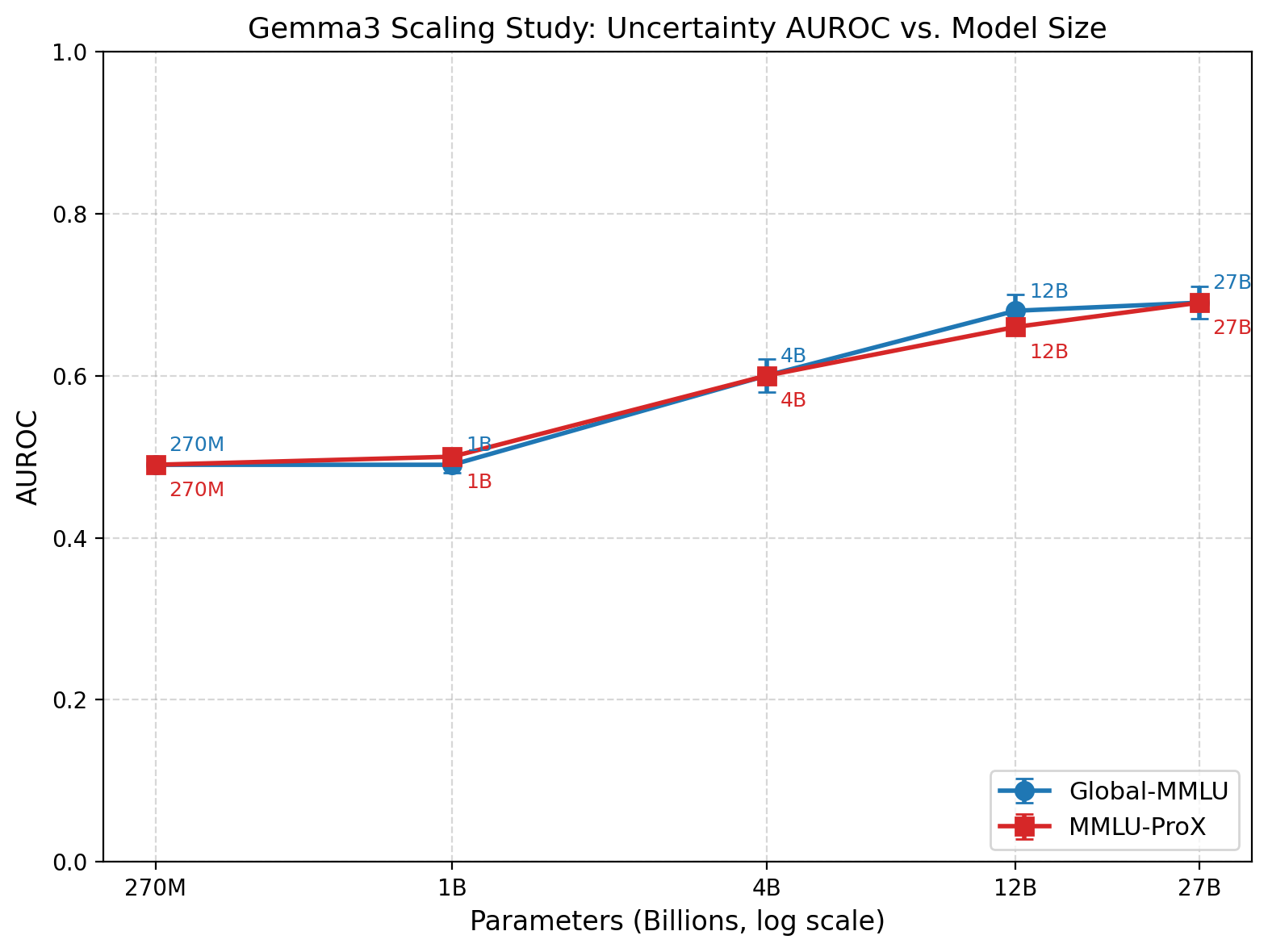}
    \caption{Uncertainty estimation performance (AUROC ± CI 95\%) across varying model scales. Results are averaged across 22 languages in both datasets. Increasing model size generally yields higher AUROC scores.}
    \label{fig:ue-method-scaling-avg}
\end{figure}

Furthermore, in Figure~\ref{fig:ue-method-scaling-avg} we show results across Gemma from 270M to 27B parameters, averaged across all nine UE methods. A clear scaling trend emerges again: models at or below 1B parameters (Gemma3-0.27B, Gemma3-1B) perform at the random baseline (0.50 AUROC), indicating that UE methods fail to extract meaningful uncertainty signals from very small models. Performance increases substantially between 1B and 4B parameters, reaching $\sim$0.60--0.65 AUROC. Beyond 12B parameters, performance plateaus around 0.67--0.70 AUROC.

We report Expected Calibration Error in Appendix~\ref{app:calibration}, where larger models show a slight tendency toward overconfidence.

\begin{figure}
    \centering
    \includegraphics[width=0.87\linewidth]{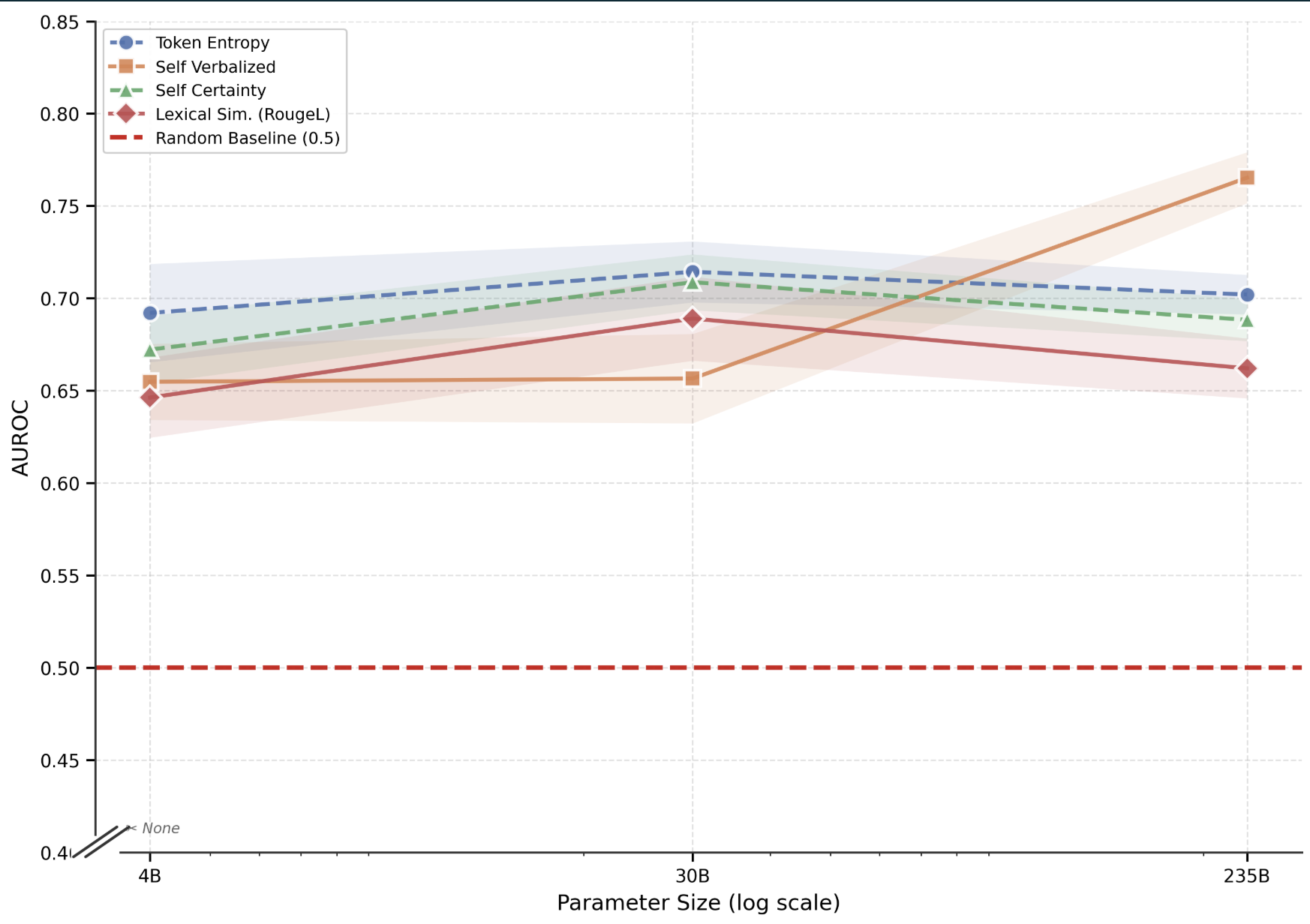}
    \caption{Effect of model scale on UE performance (Qwen3 family). Shaded regions denote 95\% CIs. Dashed line: random baseline.}
    \label{fig:ue-method-scaling}
\end{figure}

\paragraph{RQ3: How does the language of reasoning affect UE quality across languages?}

\begin{figure*}[t]
    \centering
    \includegraphics[width=0.90\linewidth]{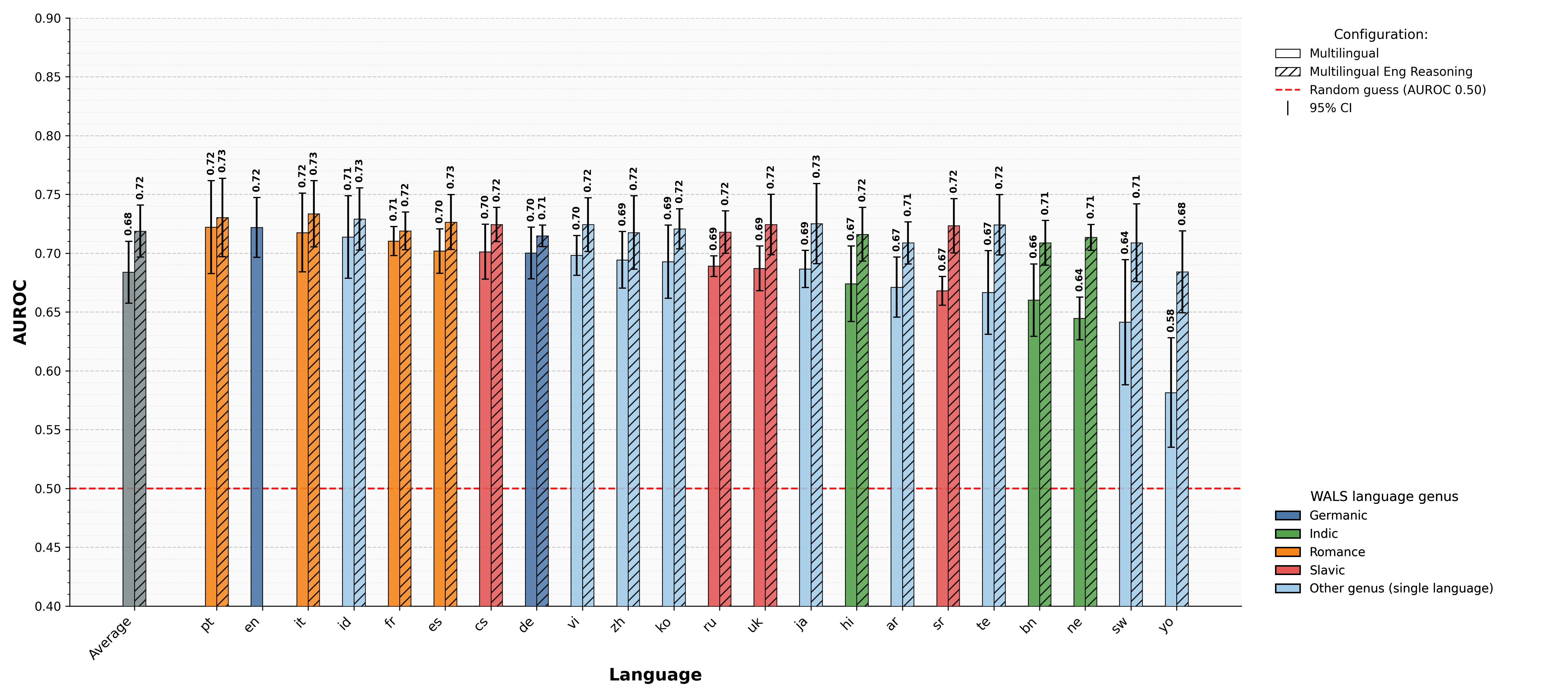}
    \caption{Effect of reasoning language on UE quality across 22 languages. 
    Solid bars represent reasoning in target language condition; hatched bars represent the English-reasoning condition. AUROC scores are averaged across all UE methods, datasets, and the largest models (Qwen3-235B, Gemma3-27B, Claude Sonnet 4.5). Error bars denote 95\% CI. English reasoning significantly narrows the performance gap for South Asian and African languages.}
\label{fig:family_barplot}
\end{figure*}

To investigate whether the language used for reasoning influences multilingual UE, we compare two configurations: \textit{Multilingual}, where the model reasons in the target language (i.e., language of the question and answers), and \textit{English Reasoning}, where reasoning is conducted in English regardless of the target language. Figure~\ref{fig:family_barplot} reports AUROC with 95\% confidence intervals across 22 languages, averaged across all datasets and the largest models (Qwen3-235B, Gemma3-27B, and Claude Sonnet 4.5), grouped by WALS language genus\footnote{\url{https://wals.info/}}.

On average, English reasoning improves UE quality (0.72 vs. 0.68 AUROC). Crucially, this improvement is driven by low-resource languages: English reasoning effectively brings their UE performance to the level of high-resource languages. For Yoruba, AUROC improves from 0.58 to 0.68; for Swahili, from 0.58 to 0.64; for Nepali, from 0.66 to 0.71—all showing non-overlapping 95\% confidence intervals. 
After applying English reasoning, these languages achieve AUROC scores comparable to Germanic and Romance languages (0.71–0.73), which show no significant change between conditions.
These results reveal an asymmetry between natural language understanding and generation. In both conditions, the question remains in the target language—only the reasoning language changes. If comprehension were the bottleneck, changing the reasoning language would not help: the model would not know what to reason about.
The fact that English reasoning closes the gap suggests that models comprehend low-resource questions adequately but struggle to generate coherent reasoning in those languages. By shifting generation to English, models leverage their stronger generative capabilities, producing stronger uncertainty signals.

Two controls confirm this interpretation:
First, some sampling-based estimators may themselves exhibit language sensitivity, as they rely on similarity (e.g., Jaccard index).
To isolate the reasoning-language effect from this potential measurement bias, we repeat the analysis using only Self Verbalized uncertainty, which relies directly on the model's reported confidence and involves no similarity scoring: the improvement under English reasoning remains significant (AUROC 0.665 $\rightarrow$ 0.679, paired bootstrap $p < 0.001$).

Second, the effect persists on the subset of questions answered identically under both conditions (88.4\% of aligned questions), where accuracy is constant by construction (AUROC 0.670 $\rightarrow$ 0.684, $p < 0.001$); the improvement in uncertainty quality is thus independent of the accuracy gains that English reasoning also delivers.
Together, these controls isolate the effect from both scoring artifacts and correctness shifts, directly supporting the generation-bottleneck interpretation.

To control for benchmark contamination, we exploit the English-reasoning condition as a memorization test: a memorized answer would be recalled regardless of the reasoning language, since question and options stay in the target language. Instead, English reasoning significantly boosts low-resource accuracy across all three model families (paired McNemar tests, FDR-corrected; Appendix~\ref{app:contamination}).

\paragraph{RQ4: How do UE methods perform under cross-lingual answer settings?}
Multilingual applications often involve cross-lingual scenarios where different components of an input span multiple languages—a common situation in cross-lingual retrieval-augmented generation (RAG) systems where documents retrieved from multiple languages must be integrated. We evaluate whether UE methods remain robust when this cross-lingual complexity is introduced. We construct instances where the question and reasoning are in the target language, but each answer option is drawn from a different language. Leveraging the parallel nature of our multilingual datasets, we use aligned answer IDs across translations and sample each option from a different language variant. For instance, an English question asking ``What day comes after Monday?'' would present answer options such as ``Mardi'' (French, Tuesday), ``Mercoledì'' (Italian, Wednesday), ``Donnerstag'' (German, Thursday), and ``Viernes'' (Spanish, Friday). Crucially, unlike RQ3, the reasoning language always follows the question language in both conditions. We evaluate using the largest models in our study: Gemma3-27B, Qwen3-235B-A22B, and Claude Sonnet 4.5.
\begin{figure*}[ht!]
    \centering
    \includegraphics[width=0.90\linewidth]{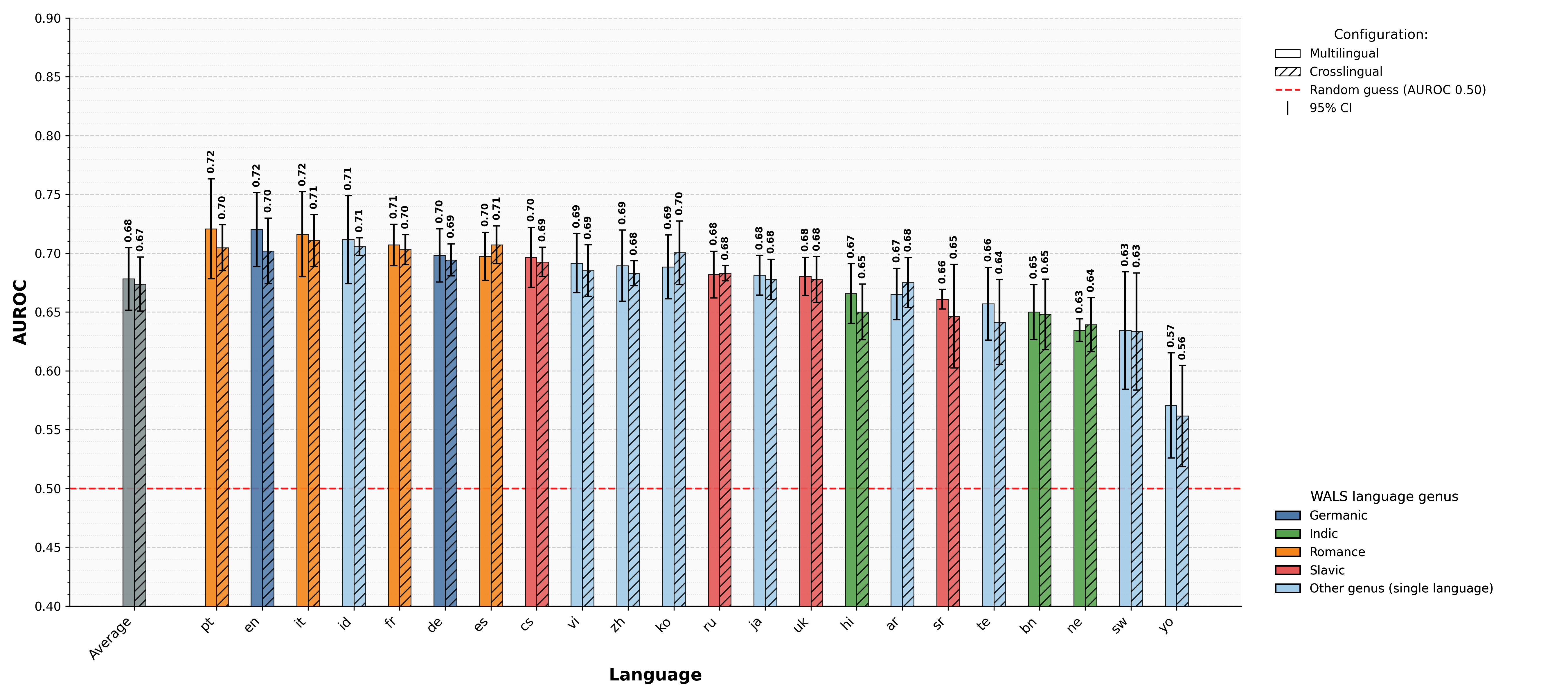}
    \caption{Effect of cross-lingual answer options on UE quality. Solid bars: multilingual setting; hatched bars: cross-lingual with each answer in a different language from the question language. Reasoning follows the question language in both conditions. AUROC scores are averaged across all UE methods, datasets, and the largest models (Qwen3-235B, Gemma3-27B, Sonnet 4.5). Error bars denote 95\% CIs.}
    \label{fig:crosslingual-answers}
\end{figure*}
Figure~\ref{fig:crosslingual-answers} reports AUROC with 95\% confidence intervals comparing the standard multilingual setting (solid bars) against the cross-lingual setting (hatched bars). This setting represents a more challenging evaluation scenario than standard multilingual UE. Despite this increased difficulty, the cross-lingual setting does not degrade UE performance. On average, AUROC remains stable (0.68 vs. 0.67) with overlapping confidence intervals across all language families. These results demonstrate that UE methods are robust to cross-lingual answer configurations, suggesting they can be reliably deployed in multilingual systems where language mixing is common.

\paragraph{RQ5: How can we find the best UE threshold across languages?}
\label{sec:rq5-selective}
A key practical application of uncertainty estimation is selective prediction: using uncertainty scores to abstain from unreliable outputs \citep{madhusudhan-etal-2025-llms}, thereby improving system trustworthiness. The central challenge in multilingual settings is threshold calibration—selecting a filtering threshold on validation data for application at inference time.
Given an uncertainty score $u$ and threshold $t$, we accept the prediction if $u < t$, otherwise abstain (e.g., defer to human review). We select $t$ by maximizing F1-score for error detection on validation data, comparing three calibration strategies: (1) $t_{\text{EN}}$, optimized on English data only and applied to all languages; (2) $t_{\text{GLOBAL}}$, optimized on pooled multilingual data; and (3) $t_{\text{LANG}}$, using language-specific thresholds transferred to corresponding target languages. We use one dataset as validation set for threshold tuning and the other as test set with Claude Sonnet 4.5, comparing against Oracle upper bounds where thresholds are tuned directly on target test data.

\begin{table}[ht!]
\centering
\resizebox{0.80\linewidth}{!}{%
\begin{tabular}{@{}lccccc@{}}
\toprule
\textbf{Strategy} &\textbf{Acc $\uparrow$} & \textbf{TP $\uparrow$} & \textbf{FP$\downarrow$} & \textbf{TN$\uparrow$} & \textbf{FN $\downarrow$} \\
\midrule
No filtering & 86.5 & 86.5 & 13.5 & 0.0 & 0.0 \\
$t_{\text{EN}}$ & 91.2  & 79.9 & 7.7 & 5.8 & 6.6 \\
$t_{\text{GLOBAL}}$ & 92.4   & 73.0 & 6.0 & 7.5 & 13.5 \\
$t_{\text{LANG}}$ & 93.0 & 72.1 & 5.4 & 8.1 & 14.4 \\
\midrule
Oracle & 93.4 & 72.2 & 5.1 & 8.4 & 14.4 \\
\bottomrule
\end{tabular}%
}
\caption{Impact of different threshold selection strategies using
Self-Verbalized (Claude Sonnet 4.5).
TP/FP: accepted predictions that are correct/wrong. TN/FN: filtered predictions that are wrong/correct. Acc is the accuracy over accepted predictions, $\text{TP}/(\text{TP}+\text{FP})$.
All values are percentages of the 44,000 samples, averaged across the two datasets.}
\label{tab:selective_prediction2}
\end{table}

Table~\ref{tab:selective_prediction2} reports the confusion matrix for each strategy. The selective prediction strategies reveal distinct trade-offs between error reduction, coverage, and deployment complexity. English-only calibration ($t_{\text{EN}}$) offers the most practical entry point, requiring minimal data collection while achieving a 43\% relative error reduction (from 13.5\% to 7.7\%) and maintaining high coverage with only 6.6\% of correct predictions filtered out. Pooled multilingual optimization ($t_{\text{GLOBAL}}$) strikes a middle ground, delivering 56\% error reduction at the cost of moderate coverage loss (13.5\% false negatives), making it suitable for applications where validation data spans multiple languages. Language-specific calibration ($t_{\text{LANG}}$) achieves the strongest practical performance with 60\% error reduction and 8.1\% error detection, nearly matching Oracle thresholds tuned on target data itself (93.0 vs. 93.4 accuracy), though this comes at the expense of filtering out 14.4\% of correct predictions. The Oracle strategy establishes the performance ceiling at 62\% error reduction but requires target-specific calibration data, limiting its practical applicability. These results demonstrate that the choice of strategy depends critically on application requirements: safety-critical systems benefit most from $t_{\text{LANG}}$'s superior error detection, resource-constrained deployments favor $t_{\text{EN}}$'s simplicity, and balanced applications achieve optimal accuracy-coverage trade-offs with $t_{\text{GLOBAL}}$.

\section{Conclusion \& Future Work}
We present the largest multilingual evaluation of LLM uncertainty estimation to date, covering 9 UE methods, 9 models (5 Gemma3, 3 Qwen3, and Claude Sonnet 4.5) spanning 270M to 235B parameters, and 22 languages from high- to low-resource settings. 

Our experiments yield three actionable findings. First, generation language matters more than question language: prompting models to reason in English closes the UE performance gap for low-resource languages, indicating that the reliability bottleneck lies in generation rather than comprehension. Notably, Swahili, Yoruba, and Telugu—the lowest-performing languages in our benchmark—achieve performance on par with English when reasoning is elicited in English. Nevertheless, reasoning in the target language is important under certain scenarios (e.g., explanations in target language for non-English speakers), and closing the gap is left for future work.

Second, UE method selection should be scale-aware. All UE methods maintain stable performance across model scales, except for Self Verbalized, which improves at scale (0.65 AUROC at 4B to 0.77 at 235B), becoming the best-performing approach with larger LLMs. This trend is consistent with prior evidence that meta-cognitive capabilities emerge primarily in larger models \citep{tian-etal-2023-just, xiongcan, steyvers2026metacognition}, although training-pipeline differences across sizes (e.g., distillation) may also contribute.

Third, threshold calibration transfers across datasets but benefits from language awareness. English-only calibration reduces error rates by 43\%, pooled multilingual calibration by 56\%, and language-specific calibration by up to 60\%. We also observe that sampling-based consistency methods degrade on low-resource languages. Since these methods rely on lexical overlap measures, sensitivity to surface-level variation may explain this degradation, though further investigation is needed.
Our evaluation framework provides a basis for further research on trustworthy multilingual AI. Future directions include extending to open-ended generation with reliable correctness assessment and developing UE methods explicitly designed for cross-linguistic robustness.

\section{Limitations}
We focus on two MCQA datasets, which, while providing reliable ground-truth labels without model-based proxies, may not fully represent the diversity of real-world NLP applications. The landscape of uncertainty estimation methods is vast, and we considered in our analysis 9 state-of-the-art methods selected to cover the main paradigms (probability-based, verbalized, and consistency-based), we did not include any training-based methods. Finally, since our benchmarks derive from MMLU, some high-resource items may have been seen during pretraining; while our English-reasoning control (Appendix~\ref{app:contamination}) argues against memorization on low-resource tiers, cross-lingual comparisons involving high-resource languages should be interpreted with this caveat in mind.

\bibliography{custom}

\clearpage
\appendix
\section{Uncertainty Estimation Methods}
\label{sec:ue-methods}

We classify Uncertainty Estimation (UE) methods into two primary categories based on the level of access required to the underlying Large Language Model (LLM). \textbf{Open-box methods} derive uncertainty from the model's internal probability distributions (logits), whereas \textbf{Closed-box methods} operate solely on the generated text, utilizing either self-reflection or the consistency between multiple stochastic samples.

Throughout, we denote the input prompt by $x$, the model's parameters by $\phi$, and the generated sequence by $\mathbf{y} = (y_1, \dots, y_T)$. The model's predictive distribution at step $t$ is denoted as $p_{\phi}(y_t \mid \mathbf{y}_{<t}, x)$.

\subsection{Open-Box Methods}
These methods require white-box access to token-level probabilities. They are computationally efficient as they typically operate on a single greedy generation, but they are inapplicable to API-based models where logits are unavailable.

\paragraph{Maximum Sequence Probability}
Following the implementation in LM-Polygraph, we estimate sequence-level
uncertainty as the negative log-probability of the generated sequence:
\begin{equation*}
    \tilde{u}_{\text{MSP}}(x)
    =
    - \sum_{t=1}^{T}
    \log p_{\phi}(y_t \mid \mathbf{y}_{<t}, x).
\end{equation*}
We then apply dataset-level min--max normalization:
\begin{equation*}
    u_{\text{MSP}}(x)
    =
    \mathrm{Norm}\left(\tilde{u}_{\text{MSP}}(x)\right).
\end{equation*}
Higher values indicate lower sequence probability and therefore greater
uncertainty.

\paragraph{Self Certainty}
Following \citet{kang2025scalable}, we measure how far each
token's predictive distribution is from a uniform distribution over the
vocabulary $\mathcal{V}$, via the KL divergence
$\mathrm{KL}(\mathcal{U}\,\|\,p_{\phi})$. Averaging over the generated tokens
and negating so that larger values denote higher uncertainty:
 \begin{equation*}
  \begin{aligned}
      u_{\text{SC}}(x) &= \mathrm{Norm}\!\left(
      \frac{1}{T}\sum_{t=1}^{T} s_t \right), \\
      s_t &= \log |\mathcal{V}|
      + \frac{1}{|\mathcal{V}|}\sum_{v\in\mathcal{V}} \log p_{\phi}(v),
  \end{aligned}
  \end{equation*}
where $p_{\phi}(v)$ is shorthand for $p_{\phi}(v \mid \mathbf{y}_{<t}, x)$. A
flatter (more uniform) predictive distribution yields higher uncertainty.

\subsection{Closed-Box Methods}
These methods treat the LLM as a black box. We further divide them into \textit{Verbalized} methods (single generation, self-reflection) and \textit{Consistency} methods (multiple generations, aggregation).

\subsubsection{Self-Verbalize}
This approach prompts the model to explicitly report its confidence, relying
on the assumption that instruction-tuned models possess internal calibration
capabilities \citep{kadavath2022language, tian-etal-2023-just}. We append a
follow-up query asking the model to report the probability $s \in [0, 1]$
that its answer is correct, and define the uncertainty as the complement of
the reported confidence:
\begin{equation*}
    u_{\text{SelfVerb}}(x) = 1 - s_{\phi}(x).
\end{equation*}
If the model fails to output a valid numerical score, we assign maximum
uncertainty ($u = 1$).

\subsubsection{Consistency and Graph-based Uncertainty}
These methods approximate uncertainty by measuring the semantic dispersion of $k$ stochastic samples $\mathcal{Y} = \{\mathbf{y}^{(1)}, \dots, \mathbf{y}^{(k)}\}$ generated at temperature $T=1$.

\paragraph{Semantic Entropy (Jaccard)}
We use a lexical variant of Semantic Entropy
\citep{DBLP:conf/iclr/KuhnGF23} in which semantic-equivalence clustering is
replaced by Jaccard-based clustering. Given $k$ stochastic generations
$\mathcal{Y}=\{\mathbf{y}^{(1)},\ldots,\mathbf{y}^{(k)}\}$, we group
generations into clusters $C_1,\ldots,C_m$ according to their Jaccard
similarity. We estimate the probability mass of each cluster empirically as

\begin{equation*}
    \hat{p}(C_j)
    =
    \frac{|C_j|}{k}.
\end{equation*}

Uncertainty is then computed using the Semantic Entropy formulation over the
resulting cluster distribution:

\begin{equation*}
    u_{\text{SemEnt}}(x)
    =
    \mathrm{Norm}\left(
        -\sum_{j=1}^{m}
        \hat{p}(C_j)
        \log \hat{p}(C_j)
    \right).
\end{equation*}

Thus, the entropy formulation follows Semantic Entropy, while Jaccard
similarity replaces entailment-based semantic equivalence when constructing
the clusters.

\paragraph{Lexical Similarity.}
We compute the average pairwise similarity using the ROUGE-L \citep{lin-2004-rouge} metric as a proxy for consistency. Higher disagreement implies higher uncertainty:
\begin{equation*}
S_{\text{LexSim}}(x)
=
\frac{1}{\binom{k}{2}}
\sum_{i<j}
\mathrm{RougeL}
\bigl(\mathbf{y}^{(i)}, \mathbf{y}^{(j)}\bigr).
\end{equation*}

\begin{equation*}
u_{\text{LexSim}}(x)
=
\mathrm{Norm}\bigl(1-S_{\text{LexSim}}(x)\bigr).
\end{equation*}

\paragraph{Graph-Theoretic Measures.}
Following \citet{lin2024generating} and the LM-Polygraph implementation, we
construct a weighted similarity matrix $W \in \mathbb{R}^{k \times k}$ over
the $k$ sampled generations. For the Jaccard variants used in our experiments,
$W_{ij}$ is the pairwise Jaccard similarity between generations
$\mathbf{y}^{(i)}$ and $\mathbf{y}^{(j)}$.

We define the degree matrix as
\begin{equation*}
    D
    =
    \operatorname{diag}
    \left(
    \sum_j W_{1j},
    \ldots,
    \sum_j W_{kj}
    \right),
\end{equation*}
and the normalized graph Laplacian as
\begin{equation*}
    L
    =
    I - D^{-1/2} W D^{-1/2}.
\end{equation*}

\begin{enumerate}

\item \textbf{Laplacian Eigenvalue (EigValLaplacian).}
Let $\lambda_1,\ldots,\lambda_k$ denote the eigenvalues of $L$.
Following LM-Polygraph, the raw uncertainty score is
\begin{equation*}
    \tilde{u}_{\text{Eig}}(x)
    =
    \sum_{r=1}^{k}
    \max\left(0,\,1-\lambda_r\right).
\end{equation*}
We report its normalized value
\begin{equation*}
    u_{\text{Eig}}(x)
    =
    \mathrm{Norm}\left(
    \tilde{u}_{\text{Eig}}(x)
    \right).
\end{equation*}

\item \textbf{Degree Matrix (DegMat).}
Following LM-Polygraph, the raw uncertainty score is
\begin{equation*}
    \tilde{u}_{\text{Deg}}(x)
    =
    \frac{
    \operatorname{tr}\left(kI-D\right)
    }{k^2}.
\end{equation*}
We report
\begin{equation*}
    u_{\text{Deg}}(x)
    =
    \mathrm{Norm}\left(
    \tilde{u}_{\text{Deg}}(x)
    \right).
\end{equation*}

\item \textbf{Eccentricity.}
Let $(\lambda_r,\mathbf{v}_r)$ denote the eigenvalue--eigenvector pairs of
the normalized graph Laplacian $L$. Following the default LM-Polygraph
implementation, we retain eigenvectors associated with eigenvalues below
the threshold $\tau_{\mathrm{eig}}=0.9$:
\begin{equation*}
    \mathcal{I}
    =
    \left\{
    r : \lambda_r < 0.9
    \right\}.
\end{equation*}

For each retained eigenvector, we subtract its mean component:
\begin{equation*}
    \bar{v}_r
    =
    \frac{1}{k}
    \sum_{i=1}^{k}
    (\mathbf{v}_r)_i,
\end{equation*}

\begin{equation*}
    \tilde{\mathbf{v}}_r
    =
    \mathbf{v}_r
    -
    \bar{v}_r \mathbf{1}.
\end{equation*}

LM-Polygraph computes
\begin{equation*}
    c_r
    =
    -
    \left\|
    \tilde{\mathbf{v}}_r
    \right\|_2,
\end{equation*}
and aggregates these components as
\begin{equation*}
    \tilde{u}_{\text{Ecc}}(x)
    =
    \left(
    \sum_{r\in\mathcal{I}}
    c_r^2
    \right)^{1/2}.
\end{equation*}

We report the normalized uncertainty
\begin{equation*}
    u_{\text{Ecc}}(x)
    =
    \mathrm{Norm}\left(
    \tilde{u}_{\text{Ecc}}(x)
    \right).
\end{equation*}

\end{enumerate}

\section{Why not open-ended QA?}
\label{app:why-not-open-ended}

An alternative to MCQA would be the use of open-ended QA datasets such as TriviaQA~\citep{joshi-etal-2017-triviaqa}, Natural~Questions~\citep{kwiatkowski-etal-2019-natural}, or their multilingual counterpart MKQA~\citep{longpre-etal-2021-mkqa}.
We do not adopt this setup, because correctness on these datasets must still be assessed on the open-ended string, which requires regex matching, an LLM-as-a-judge, or embedding similarity, reintroducing exactly the biases in the correctness signal that we want to avoid.
Open-ended answers are also extremely short factoids (e.g.\ ``Albert Einstein''): the median is $1$ word for TriviaQA and $2$ words for Natural Questions, so they do not provide longer text for UE than MCQA labels do, and reasoning elicitation would be needed in either case.

MCQA with elicited reasoning is therefore the only configuration that simultaneously satisfies all three requirements: (1) multilingual coverage including low-resource languages, (2) sufficiently long generations for UE, (3) and unbiased model-free correctness.

\section{Languages}
\label{appendix-languages}
Our benchmark covers 22 languages given by the intersection of Global-MMLU and MMLU-ProX, so that every instance has a strictly parallel multiple-choice question and shared answer key across all languages.

The languages and their ISO codes are:
Arabic (ar), Bengali (bn), Czech (cs), German (de), English (en), Spanish (es), French (fr), Hindi (hi), Indonesian (id), Italian (it), Japanese (ja), Korean (ko), Nepali (ne), Portuguese (pt), Russian (ru), Serbian (sr), Swahili (sw), Telugu (te), Ukrainian (uk), Vietnamese (vi), Yoruba (yo), and Chinese (zh).

\section{Prompts}\label{app:prompts}
This section details the prompts used in our experiments. All uncertainty estimation methods share the same base prompt for answer generation. The \textit{self-verbalized} method additionally employs a follow-up prompt to elicit a confidence score from the model.

\subsection{Answer Generation Prompt}
\begin{lstlisting}[basicstyle=\ttfamily\small, breaklines=true, frame=single]
Provide your best guess for the following question about {subject} selecting one of the options. Before giving your answer, provide a step-by-step explanation of your thought process. Then on a new line give the guess with no other words or explanation.

For example:

Explanation: <one sentence step-by-step explanation of your thought process>
Guess: <most likely guess, as short as possible; not a complete sentence, just the guess!>

Q: {question}
A. {choice_A}
B. {choice_B}
C. {choice_C}
D. {choice_D}
\end{lstlisting}

\subsection{Self-Verbalized Confidence Prompt}
\begin{lstlisting}[basicstyle=\ttfamily\small, breaklines=true, frame=single]
Provide the probability that your guess is correct. Give ONLY the probability, no other words or explanation.

For example:

Probability: <the probability between 0.0 and 1.0 that your guess is correct, without any extra commentary whatsoever; just the probability!>
\end{lstlisting}

To transform the probability into the uncertainty score we computed the uncertainty as $1 - $probability score.

\subsection{Prompt Phrasing Ablation for Self-Verbalized Confidence}
\label{app:prompt-ablation}
We re-ran the Self-Verbalized method with Qwen3-235B on 200 English questions
(100 from Global-MMLU, 100 from MMLU-ProX) under three elicitation phrasings.
Error-detection AUROC was stable across wordings (Variant 1: 0.72; Variant 2:
0.74; Variant 3: 0.76), with no statistically significant difference between
them.

\subsubsection{Variant 1 --- Baseline.}
\begin{lstlisting}[basicstyle=\ttfamily\small, breaklines=true, frame=single]
The following is a multiple-choice question. Think step by step, then give your final answer and your confidence.

Follow this format exactly:
<rationale> [your step-by-step reasoning] </rationale>
<final> [the answer letter only, e.g. A] </final>
<verbalized_confidence> [confidence score between 0 and 100] </verbalized_confidence>
\end{lstlisting}

\subsubsection{Variant 2 --- ``How sure are you''.}
\begin{lstlisting}[basicstyle=\ttfamily\small, breaklines=true, frame=single]
Answer the following multiple-choice question. First reason through it, then state which option is correct and how sure you are on a 0 to 100 scale.

Respond strictly in this format:
<rationale> [your reasoning] </rationale>
<final> [single answer letter] </final>
<verbalized_confidence> [an integer from 0 to 100 for how confident you are] </verbalized_confidence>
\end{lstlisting}

\subsubsection{Variant 3 --- ``Probability correct''.}
\begin{lstlisting}[basicstyle=\ttfamily\small, breaklines=true, frame=single]
You are answering a multiple-choice question. Work through it, choose the best option, and estimate the probability that your answer is correct.

Use exactly this format:
<rationale> [reasoning steps] </rationale>
<final> [answer letter] </final>
<verbalized_confidence> [a number 0-100 giving the percent chance your answer is correct] </verbalized_confidence>
\end{lstlisting}

\section{Question Answering Categories}\label{app:question-answering-categories}
We report the list of categories in both datasets used.
The categories of Global-MMLU are [business, humanities, medical, other, stem, social sciences].
The categories of MMLU-ProX are [biology, business, chemistry, computer science, economics, engineering, health, history, law, math, other, philosophy, physics, psychology]

\section{Cross-lingual Q\&A Accuracy comparison}
This section provides task accuracy results for the cross-lingual evaluation discussed in RQ4 (Section~\ref{sec:results}). In the cross-lingual setting, the question and reasoning are in the target language, but each answer option is drawn from a different language.

We compared the task accuracy between the standard multilingual setting and the cross-lingual setting. As expected, the cross-lingual setting is more challenging: accuracy drops by 4--6 percentage points across all models across all the scales. This decrease reflects the additional difficulty of matching reasoning to answer options expressed in different languages.

Crucially, while task accuracy decreases, uncertainty estimation performance (AUROC) remains stable (see Figure~\ref{fig:crosslingual-answers}). This indicates that UE methods successfully capture model confidence regardless of the cross-lingual configuration, and do not rely solely on lexical matching between reasoning and answer options.

Beyond uncertainty estimation, we also investigate whether English reasoning
improves task accuracy. Table~\ref{tab:eng-reasoning-accuracy} compares
language-specific and English reasoning in terms of Q\&A accuracy, showing
the relative improvement ($\Delta_{\text{rel}}\%$) and McNemar significance
across the 21 non-English languages, pooled across Gemma3-27B, Qwen3-30B,
and Claude Sonnet 4.5 on Global-MMLU and MMLU-ProX.

The results indicate that English reasoning consistently improves task
accuracy, with an average relative improvement of $+3.3\%$ and statistically
significant gains in 19 of 21 languages. The improvement is most pronounced
for lower-resource languages such as Yoruba (YO, $+12.8\%$), Swahili (SW,
$+8.8\%$), and Telugu (TE, $+7.8\%$), while higher-resource languages such as
Italian (IT), Vietnamese (VI), and Portuguese (PT) exhibit smaller gains.
This suggests that English reasoning not only helps models better recognize
their uncertainty but also enhances their ability to answer correctly,
particularly in languages where the model's native reasoning capabilities are
weaker.

\begin{table}[h]
    \centering
    \small
    \begin{tabular}{lcccc}
        \toprule
        \textbf{Lang.} & \textbf{Lang-Spec.} & \textbf{Eng. Reason.} & \textbf{$\Delta_{\text{rel}}\%$} & \textbf{Sig} \\
        \midrule
        YO & 0.457 & 0.516 & +12.8\% & *** \\
        SW & 0.588 & 0.639 & +8.8\%  & *** \\
        TE & 0.691 & 0.745 & +7.8\%  & *** \\
        NE & 0.705 & 0.750 & +6.4\%  & *** \\
        BN & 0.697 & 0.738 & +5.8\%  & *** \\
        HI & 0.709 & 0.749 & +5.5\%  & *** \\
        AR & 0.719 & 0.741 & +3.1\%  & *** \\
        KO & 0.725 & 0.748 & +3.1\%  & *** \\
        UK & 0.745 & 0.765 & +2.6\%  & *** \\
        SR & 0.737 & 0.753 & +2.2\%  & *** \\
        CS & 0.749 & 0.765 & +2.1\%  & *** \\
        ES & 0.753 & 0.768 & +2.0\%  & *** \\
        JA & 0.733 & 0.746 & +1.8\%  & **  \\
        RU & 0.747 & 0.761 & +1.8\%  & **  \\
        FR & 0.750 & 0.762 & +1.7\%  & **  \\
        ZH & 0.740 & 0.751 & +1.5\%  & **  \\
        DE & 0.747 & 0.758 & +1.5\%  & **  \\
        ID & 0.751 & 0.761 & +1.3\%  & *   \\
        PT & 0.759 & 0.769 & +1.3\%  & **  \\
        VI & 0.750 & 0.755 & +0.7\%  & ns  \\
        IT & 0.765 & 0.768 & +0.4\%  & ns  \\
        \midrule
        \textbf{Avg} & \textbf{0.715} & \textbf{0.739} & \textbf{+3.3\%} & \\
        \bottomrule
    \end{tabular}
    \caption{Q\&A accuracy under language-specific vs.\ English reasoning,
    pooled across Gemma3-27B, Qwen3-30B, and Claude Sonnet 4.5 on Global-MMLU
    and MMLU-ProX. Paired McNemar tests per language, BH-FDR adjusted across
    the 21 non-English languages. Significance: *** $p<0.001$, ** $p<0.01$,
    * $p<0.05$, ns = not significant.}
    \label{tab:eng-reasoning-accuracy}
\end{table}

\section{Does the quantization impact multilingual UE performance?}\label{app:quantization}
\begin{figure}
    \centering
    \includegraphics[width=\linewidth]{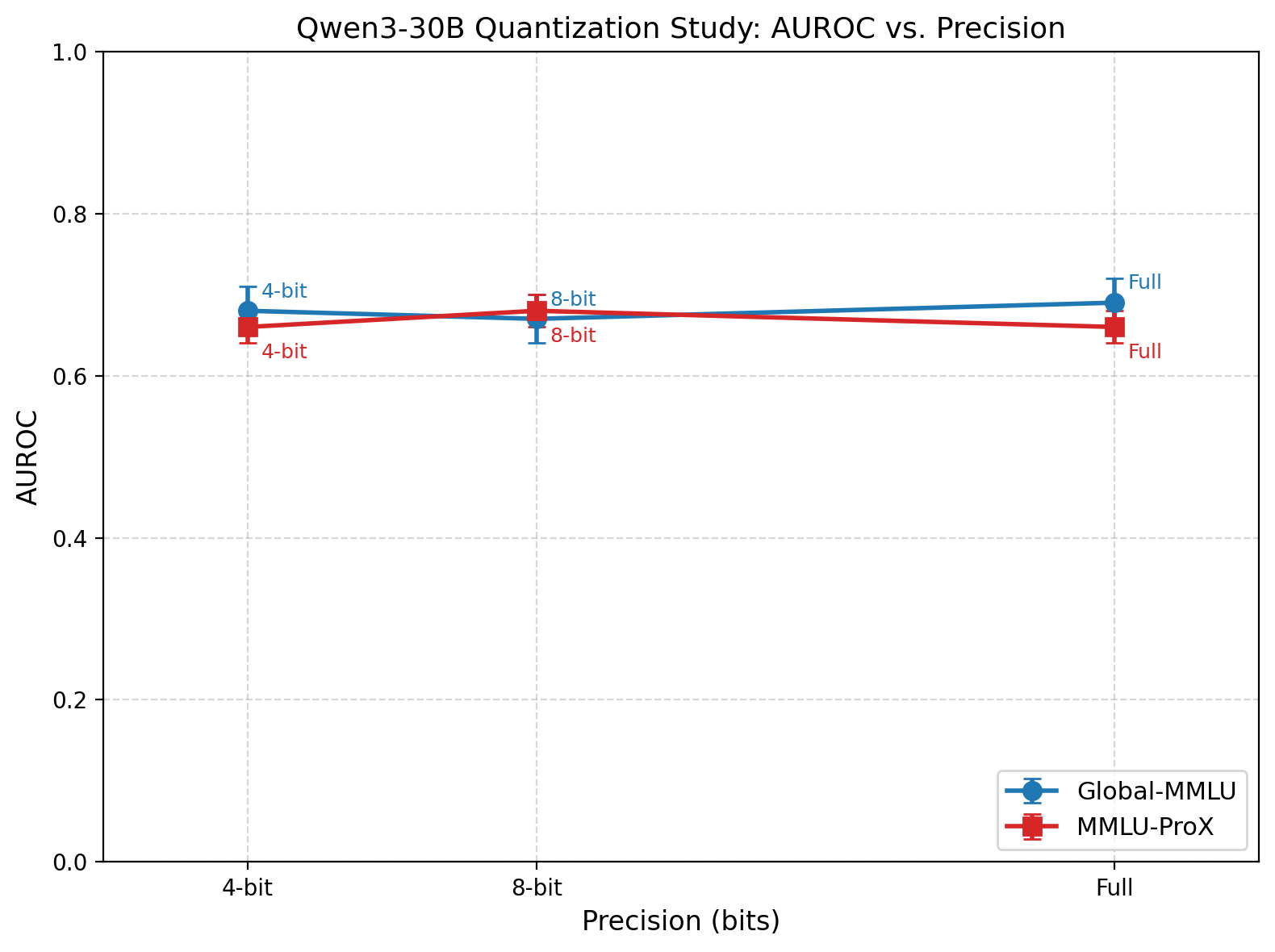}
    \caption{Uncertainty estimation performance for Qwen3-30B with quantization to 8-bit and 4-bit. Quantization has minimal impact on both accuracy and uncertainty estimation at this model scale. (± indicates 95\% CI).}
    \label{fig:quantiziation}
\end{figure}

 In Figure \ref{fig:quantiziation}, we compare the performance of full-precision Qwen3-30B against its 8-bit and 4-bit quantizations. We selected Qwen3-30B as it is the largest model in our study that fits in memory at full precision, making it the most relevant candidate for quantization. As can be noticed, there is no statistically significant impact from quantizing the model.

\section{Hardware Infrastructure}\label{app:hardware-infrastructure}
All experiments were conducted on an Amazon EC2 g6e.48xlarge instance with an AMD EPYC 7R13 processor (192 vCPUs), 1.5 TiB of system memory, and 8 NVIDIA L40S GPUs (48GB VRAM each, 384GB total).

\section{Complete Multilingual UE Method Scaling Effect}\label{app:model_scale_all_ue}
\begin{figure}
    \centering
    \includegraphics[width=\linewidth]{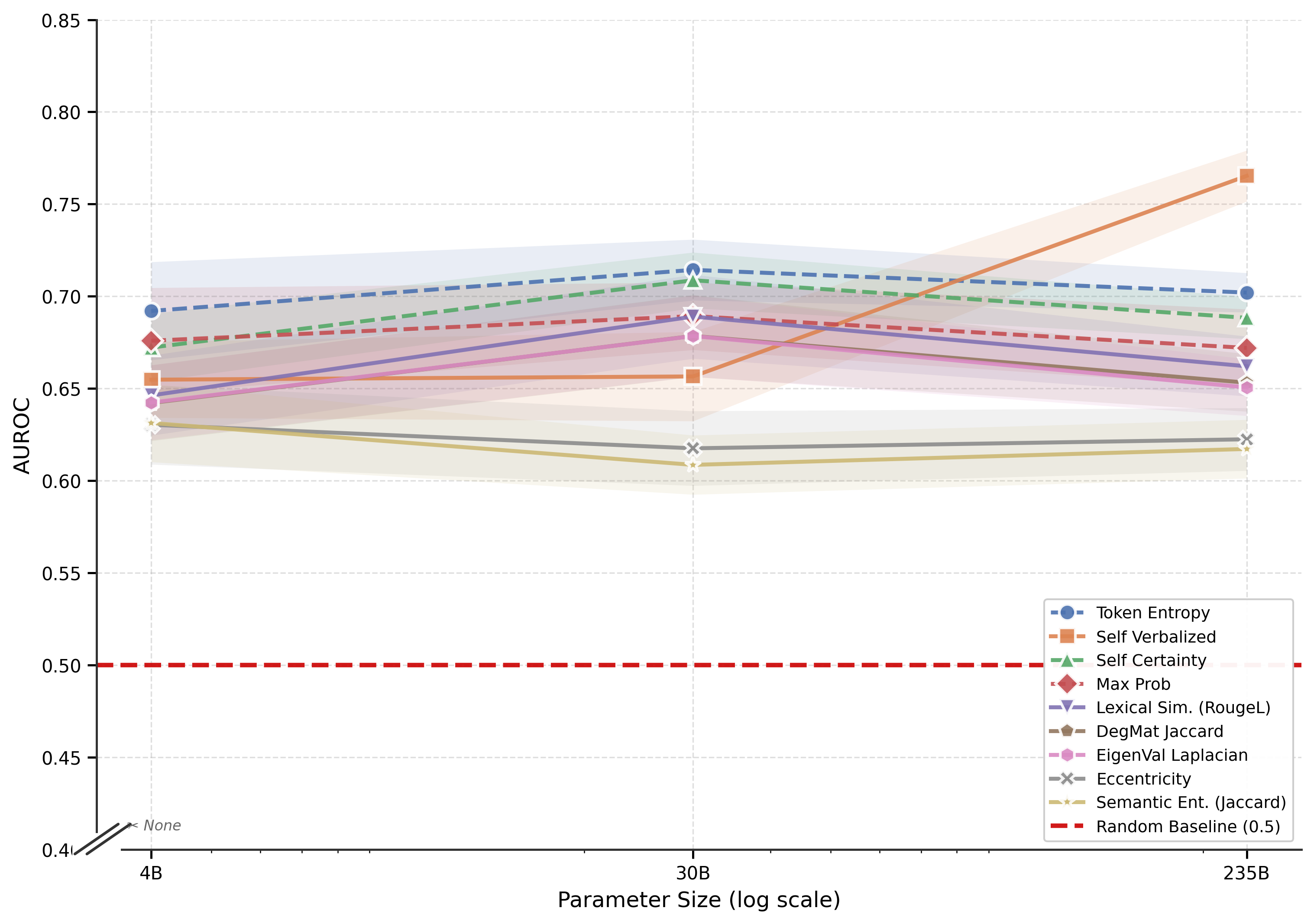}
    \caption{Effect of model scale on UE performance (Qwen3 family). Shaded regions denote 95\% CIs. Self Verbalized improves with statistical significance at 235B. Dashed line: random baseline. This figure complements Figure \ref{fig:ue-method-scaling} presented in the main where for simplicity we show only 4 methods.}
    \label{fig:ue-method-scaling-all-ue}
\end{figure}

In Figure \ref{fig:ue-method-scaling-all-ue}, we report the model size scaling factor with all the UE methods.

\section{Label Distribution Analysis for Positional Bias}
\label{app:label-distribution}

To rule out positional bias in our evaluation, we analyzed the distribution of correct answer labels across the two MCQA benchmarks used in our study. Tables~\ref{tab:global-mmlu-dist} and~\ref{tab:mmlu-prox-dist} show that, for both Global MMLU (4 options, expected uniform $=25\%$) and MMLU-ProX (10 options, expected uniform $=10\%$), all labels fall within approximately $\pm 2$ percentage points of the uniform expectation. We therefore conclude that label distributions are sufficiently balanced to exclude positional bias as a confounding factor in the results reported in the main paper.

\begin{table}[h]
    \centering
    \begin{tabular}{lcccc}
        \toprule
        \textbf{Label} & A & B & C & D \\
        \midrule
        \textbf{Freq. (\%)} & 22.96 & 24.65 & 25.50 & 26.90 \\
        \bottomrule
    \end{tabular}
    \caption{Label distribution for Global MMLU.}
    \label{tab:global-mmlu-dist}
\end{table}

\begin{table}[h]
    \centering
    \small
    \begin{tabular}{lc}
        \toprule
        \textbf{Label} & \textbf{Freq. (\%)} \\
        \midrule
        A & 11.57 \\
        B & 11.14 \\
        C & 10.93 \\
        D & 11.11 \\
        E & 9.58 \\
        F & 9.42 \\
        G & 9.78 \\
        H & 9.32 \\
        I & 9.22 \\
        J & 7.92 \\
        \bottomrule
    \end{tabular}
    \caption{Label distribution for MMLU-ProX.}
    \label{tab:mmlu-prox-dist}
\end{table}

\section{Calibration Analysis}\label{app:calibration}
\begin{figure}
    \centering
    \includegraphics[width=\linewidth]{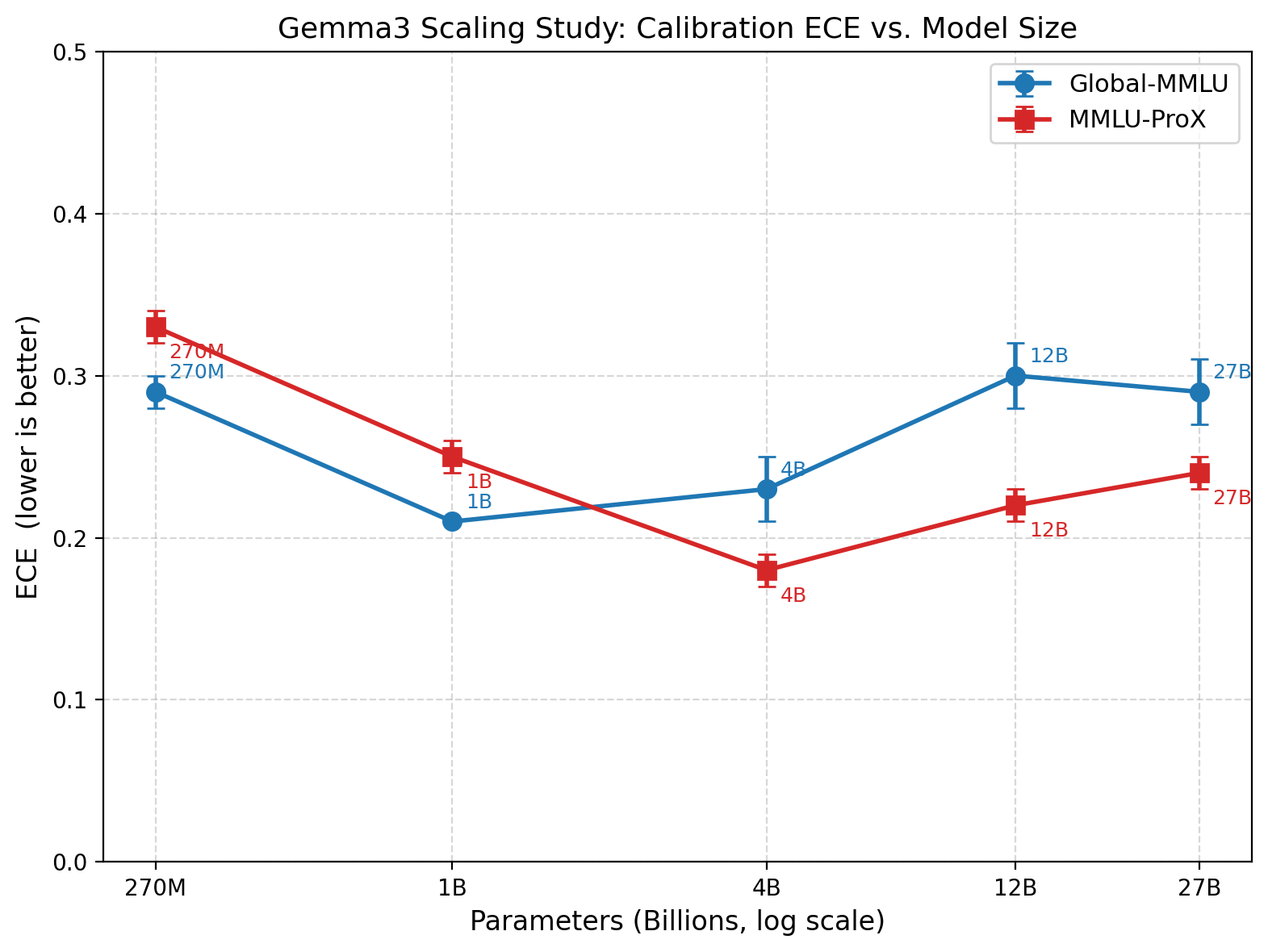}
    \caption{Effect of model scale on ECE. (Mean across langs and 95\% CIs).}
    \label{fig:ece-scaling}
\end{figure}

Figure~\ref{fig:ece-scaling} extends the AUROC results from the main paper (Figure~\ref{fig:ue-method-scaling-avg}) with Expected Calibration Error (ECE), which measures the alignment between predicted confidence and actual accuracy. Lower ECE indicates better calibration.

\section{Statistical Significance Analysis}
\label{app:significance}
To complement the 95\% confidence intervals reported in the main paper, we
perform formal hypothesis testing with the DeLong test for ROC curves,
applying Benjamini--Hochberg (BH) FDR correction across all 22 languages,
complementing the confidence intervals shown in
Figures~\ref{fig:ue-method-scaling} and~\ref{fig:ue-method-scaling-all-ue}.
We report per-language results for two analyses on Global-MMLU; note that
per-language values on a single dataset differ from the cross-dataset
averages shown in Figure~\ref{fig:ue-method-scaling}.

\paragraph{Model scaling (RQ2).} We test Self-Verbalized AUROC on Qwen3-235B
vs.\ Qwen3-30B. Because the two models produce distinct correctness labels,
their ROC curves are independent and we use an unpaired two-sided DeLong test
per language. The scaling advantage is significant in 22 of 22 languages
after FDR correction (mean gap: $+0.153$ AUROC).

\begin{table}[h]
\centering
\small
\begin{tabular}{lccccc}
\toprule
\textbf{Lang} & \textbf{235B} & \textbf{30B} & \textbf{$\Delta$} & \textbf{$p$ (BH)} & \textbf{Sig} \\
\midrule
ar & 0.693 & 0.558 & +0.135 & 3.3e-05 & *** \\
bn & 0.710 & 0.582 & +0.129 & 2.2e-05 & *** \\
cs & 0.731 & 0.623 & +0.108 & 1.3e-03 & **  \\
de & 0.725 & 0.585 & +0.139 & 1.3e-04 & *** \\
en & 0.805 & 0.571 & +0.234 & 5.0e-09 & *** \\
es & 0.728 & 0.579 & +0.149 & 6.1e-05 & *** \\
fr & 0.711 & 0.566 & +0.145 & 4.6e-05 & *** \\
hi & 0.746 & 0.621 & +0.125 & 4.6e-05 & *** \\
id & 0.742 & 0.585 & +0.157 & 6.4e-06 & *** \\
it & 0.748 & 0.578 & +0.170 & 5.0e-06 & *** \\
ja & 0.748 & 0.556 & +0.192 & 3.6e-10 & *** \\
ko & 0.746 & 0.542 & +0.204 & 3.3e-09 & *** \\
ne & 0.700 & 0.593 & +0.106 & 4.4e-04 & *** \\
pt & 0.742 & 0.567 & +0.175 & 2.3e-06 & *** \\
ru & 0.771 & 0.554 & +0.218 & 3.6e-10 & *** \\
sr & 0.744 & 0.551 & +0.192 & 3.8e-09 & *** \\
sw & 0.670 & 0.575 & +0.096 & 1.5e-04 & *** \\
te & 0.770 & 0.608 & +0.162 & 1.8e-07 & *** \\
uk & 0.771 & 0.613 & +0.159 & 4.0e-06 & *** \\
vi & 0.715 & 0.584 & +0.130 & 6.7e-05 & *** \\
yo & 0.655 & 0.549 & +0.105 & 4.6e-05 & *** \\
zh & 0.732 & 0.592 & +0.140 & 3.3e-05 & *** \\
\bottomrule
\end{tabular}
\caption{Scaling within the Qwen3 family (235B vs.\ 30B), Self-Verbalized
AUROC on Global-MMLU, all 22 languages. Unpaired DeLong test, BH-FDR
adjusted. Significance: *** $p<0.001$, ** $p<0.01$, * $p<0.05$,
ns = not significant.}
\label{tab:delong-scaling}
\end{table}

\paragraph{Method ranking (RQ1/RQ2).} We compare Self-Verbalized vs.\ Token
Entropy on Qwen3-30B (our largest unquantized open-box baseline) with a
paired DeLong test for correlated ROC curves, as both methods are scored on
identical samples under identical correctness labels. The difference is
significant in 19 of 22 languages after correction, uniformly favoring Token
Entropy across all significant outcomes (mean gap: $0.144$), consistent with
the scale-dependent reversal discussed in RQ2.

\begin{table}[h]
\centering
\small
\begin{tabular}{lccccc}
\toprule
\textbf{Lang} & \textbf{TE} & \textbf{SV} & \textbf{$\Delta$ (SV$-$TE)} & \textbf{$p$ (BH)} & \textbf{Sig} \\
\midrule
ar & 0.725 & 0.558 & $-0.167$ & 7.2e-08 & *** \\
bn & 0.645 & 0.582 & $-0.064$ & 3.0e-02 & *   \\
cs & 0.780 & 0.623 & $-0.158$ & 5.6e-08 & *** \\
de & 0.726 & 0.585 & $-0.141$ & 1.2e-05 & *** \\
en & 0.773 & 0.571 & $-0.202$ & 2.7e-07 & *** \\
es & 0.760 & 0.579 & $-0.181$ & 2.8e-07 & *** \\
fr & 0.774 & 0.566 & $-0.208$ & 4.5e-12 & *** \\
hi & 0.664 & 0.621 & $-0.043$ & 1.5e-01 & ns  \\
id & 0.789 & 0.585 & $-0.204$ & 4.5e-12 & *** \\
it & 0.811 & 0.578 & $-0.232$ & 2.6e-11 & *** \\
ja & 0.754 & 0.556 & $-0.198$ & 1.8e-10 & *** \\
ko & 0.696 & 0.542 & $-0.153$ & 1.6e-06 & *** \\
ne & 0.673 & 0.593 & $-0.080$ & 1.2e-02 & *   \\
pt & 0.795 & 0.567 & $-0.228$ & 1.7e-13 & *** \\
ru & 0.767 & 0.554 & $-0.214$ & 1.6e-11 & *** \\
sr & 0.732 & 0.551 & $-0.180$ & 1.1e-09 & *** \\
sw & 0.636 & 0.575 & $-0.061$ & 1.6e-02 & *   \\
te & 0.631 & 0.608 & $-0.022$ & 4.7e-01 & ns  \\
uk & 0.746 & 0.613 & $-0.134$ & 4.7e-05 & *** \\
vi & 0.744 & 0.584 & $-0.160$ & 1.9e-07 & *** \\
yo & 0.514 & 0.549 & $+0.035$ & 2.3e-01 & ns  \\
zh & 0.754 & 0.592 & $-0.162$ & 1.6e-06 & *** \\
\bottomrule
\end{tabular}
\caption{Method ranking (Token Entropy (TE) vs.\ Self-Verbalized) on Qwen3-30B,
AUROC on Global-MMLU, all 22 languages. Paired DeLong test, BH-FDR adjusted.
Significance: *** $p<0.001$, ** $p<0.01$, * $p<0.05$, ns = not significant.}
\label{tab:delong-methods}
\end{table}

\paragraph{Scale-dependent reversal (RQ2).} The per-language DeLong test above
compares methods within each language ($\sim$600 items).
To test the scale-dependent \emph{ranking} reversal visible in
Figure~\ref{fig:ue-method-scaling}, which concerns the mean across languages,
we use a paired Wilcoxon signed-rank test over the 22 per-language AUROC
values on Global-MMLU.
At 30B, Token Entropy significantly outperforms Self-Verbalized (mean $0.722$ vs.\ $0.579$; $W=2.0$, $p<10^{-5}$; cf.\ Table~\ref{tab:delong-methods}).
At 235B the ranking reverses: Self-Verbalized significantly outperforms Token Entropy (mean $0.732$ vs.\ $0.696$; higher in 17/22 languages; $W=31.0$, $p=1.1\times10^{-3}$), confirming the crossover.

\section{Contamination Control: English Reasoning as a Memorization Test}
\label{app:contamination}
If a model had memorized an answer during pretraining, it should recall it
regardless of the reasoning language, since the question and options remain
in the target language in both conditions. We test this with paired McNemar
tests (BH-FDR corrected across the 21 non-English languages) on the
predictions pooled across Gemma3-27B, Qwen3-30B, and Claude Sonnet 4.5 on
both datasets.
Table~\ref{tab:eng-reasoning-accuracy} reports the
per-language accuracies with the corresponding significance levels: English
reasoning significantly improves accuracy in 19 of 21 languages, with the
largest gains on the lowest-resource languages (Yoruba $+12.8\%$, Swahili
$+8.8\%$, Telugu $+7.8\%$); on these languages the improvement is also
significant within each model individually ($p<0.05$).

\end{document}